# Self-Aware Trajectory Prediction for Safe Autonomous Driving

Wenbo Shao, Jun Li, and Hong Wang*

*Abstract*—Trajectory prediction is one of the key components of the autonomous driving software stack. Accurate prediction for the future movement of surrounding traffic participants is an important prerequisite for ensuring the driving efficiency and safety of intelligent vehicles. Trajectory prediction algorithms based on artificial intelligence have been widely studied and applied in recent years and have achieved remarkable results. However, complex artificial intelligence models are uncertain and difficult to explain, so they may face unintended failures when applied in the real world. In this paper, a self-aware trajectory prediction method is proposed. By introducing a self-awareness module and a two-stage training process, the original trajectory prediction module's performance is estimated online, to facilitate the system to deal with the possible scenario of insufficient prediction function in time, and create conditions for the realization of safe and reliable autonomous driving. Comprehensive experiments and analysis are performed, and the proposed method performed well in terms of self-awareness, memory footprint, and real-time performance, showing that it may serve as a promising paradigm for safe autonomous driving.

## I. Introduction

Intelligent vehicles (IVs) play a significant role in improving road traffic and mobility, with trajectory prediction being a key area of focus [1]. Accurate prediction of the future motion of traffic participants (TPs) can enhance the efficiency and safety of IVs' decision-making. Artificial intelligence (AI) techniques, particularly deep learning, have significantly improved trajectory prediction accuracy and achieved state-of-the-art results in major autonomous driving (AD) competitions and datasets [2], [3]. However, extensive research and application on AI has also raised concerns about ensuring safety for automated driving systems such as safety of the intended functionality (SOTIF). There are inherent uncertainties and black-box problems in AI algorithms. Moreover, numerous unknown long-tail scenarios in the real world may cause unpredictable failures of trajectory predictor, it may lead to unacceptable decision-making errors resulting in serious traffic accidents [4].

*Research supported by the National Science Foundation of China Project: U1964203 and 52072215, and the National Key R&D Program of China: 2022YFB2503003 and 2020YFB1600303. (*Corresponding author: Hong Wang*)

Wenbo Shao, Jun Li and Hong Wang are with the School of Vehicle and Mobility, Tsinghua University, Beijing 100084, China. (e-mail: swb19@mails.tsinghua.edu.cn; lj19580324@126.com, hong_wang@mail.tsinghua.edu.cn)

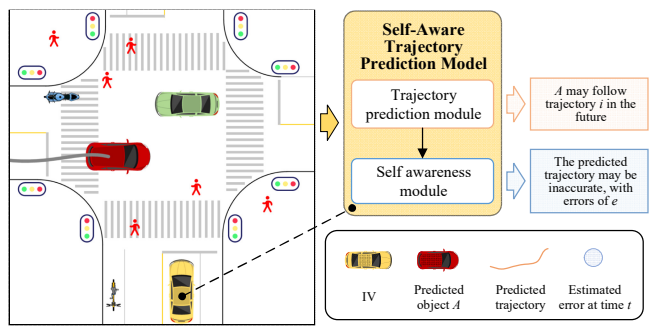

Fig. 1. Endowing the trajectory prediction model with self-awareness.

To cope with the above challenges, endowing autonomous driving systems with self-awareness is an important technical direction [5], [6]. Specifically, as shown in Fig. 1, a self-aware trajectory prediction model is constructed for IV, which can diagnose the prediction results while predicting the trajectory, thus alleviating the risk from the functional insufficiency of the original prediction module, it constitutes a potentially useful ingredient in contributing to safe, reliable, and robust IV.

In this study., we focus on self-aware trajectory prediction for safe AD. By proposing a new framework and its training process, the lack of self-awareness of the trajectory prediction module is remedied. The main contributions are as follows:

- **A novel self-aware trajectory prediction framework.** A new, computationally efficient self-awareness module is introduced to diagnose the prediction results online by obtaining real-time information from the middle layer and output layer of the original trajectory prediction module.

- **A new training and inference pipeline for the proposed framework.** We propose a two-stage training process to optimize the self-awareness module without affecting the performance of the original trajectory prediction module. The training of self-awareness module does not need to modify the parameters of the trained trajectory prediction module, but only needs to obtain part of its information, so it has strong flexibility and practicability.

- **Comprehensive experiments and analysis.** Various baselines are implemented for comparison, showing the superiority of the proposed method in terms of self-awareness, memory usage, and real-time performance. In addition, detailed ablation studies, and quantitative and qualitative analyses provide an in-depth exploration.

The remainder of this paper is organized as follows: Section II presents existing related work. Section III illustrates the proposed method. Section IV introduces the specific settings in the experiments. Section V reports and analyzes the experimental results. Section VI summarizes the work.

## II. RELATED WORK

### A. Trajectory Prediction

Early researchers generally adopted the kinematic model or dynamic model to realize trajectory prediction, such as the constant velocity, constant acceleration, and constant angular acceleration model [1], which showed robust performance in short-term prediction. In recent years, AI-based models have shown significant performance advantages. In particular, the emergence of various deep neural networks (DNNs) has provided strong support for breakthroughs in the field of trajectory prediction. Long short-term memory (LSTM)-based and gated recurrent unit (GRU)-based networks [7] provide convenience for modeling temporal relationships. Transformer [8] improves the networks' ability to predict long time series by introducing the attention mechanism. graph neural networks (GNNs) [9], such as graph convolution and graph attention, provide convenience for modeling interaction relationships.

Different algorithms have been proposed to optimize the prediction performance. Social-LSTM [7] adopts a social pooling layer to model interactions. GRIP++ [9] uses both fixed and dynamic graphs to describe interactions between different types of objects. VectorNet [10] and LaneGCN [11] adopt vector graphs to represent roads and model the actor-map interaction. TNT [12] and denseTNT [13] extract goal states to effectively capture the TPs' future motion pattern, thus generating predicted trajectories conditioned on goals. Trajectron++ [14] achieves dynamically feasible trajectory prediction by integrating the dynamic model into the networks' output part and building a differentiable structure. M2I [3] constructs conflict-aware prediction models through the in-depth exploration of conflict relations, which improves the models' ability to cope with key interactions. However, existing researches on trajectory prediction focus on the improvement of prediction performance itself but lack relevant feedback for scenarios that may cause insufficient prediction, which may lead to unacceptable risks for IVs in these scenarios. Different from the previous work, the proposed method is dedicated to endowing the trajectory prediction module with self-awareness.

### B. Introspective Artificial Intelligence

With the application of AI in safety-critical scenarios, there has been an increasing amount of research focusing on the introspection or self-awareness ability of AI models. Uncertainty has been extensively studied as a measure of how confident a model is in its output [15]. Epistemic uncertainty (EU) reflects how well a model understands the current input, and various methods have been proposed and improved to quantify the EU. Bayesian Neural Network (BNN) [16], as a method with a strong theoretical basis, has attracted attention in the early stage. However, with the increase in network complexity, BNN is difficult to be practically applied due to its high requirements for computing resources. Subsequently, Monte Carlo (MC) dropout and its variants [17], [18] were proposed as easy-to-implement approximate inference methods. Additionally, the non-Bayesian methods represented by deep ensemble [19], [20] have been widely used due to their simplicity and scalability, and have achieved good results. However, the operation of MC dropout and deep ensemble still requires multiple forward propagations of the networks to obtain the uncertainty estimation results, so their application in a system with finite computing resources is limited.

AI has a strong dependence on data, and various anomaly detection methods [21], [22] have been proposed to detect outliers from the perspective of data, including unary classification, cluster analysis, and reconstruction-based methods. Autoencoder (AE) [23] is one of the representative methods, its basic assumption is that outliers cannot be compressed and then reconstructed from latent space efficiently. In practice, some studies apply the self-awareness of AI models to the out-of-distribution (OOD) detection task [24], to identify unseen inputs during training for targeted protection. Some studies focus on error detection for AI models [25], [26] so that AI models can timely identify and discard false predictions during online inference. In this study, a separate self-awareness module is introduced to enable the trajectory prediction model to self-diagnose prediction errors.

### C. Endowing AD Systems with Self-Awareness

Constructing self-aware AD systems is beneficial for increasing system reliability and mitigating risks caused by system problems [5], [6]. In this process, how to let the systems know when and what they don't know is the key [27]. In recent years, the identification of system failures has attracted increasing attention [28]. For the perception module, mutual verification and fusion based on multi-source perception information [29], [30] or multi-subtasks [31] is an important way. In addition, the aforementioned uncertainty and anomaly detection methods [32], [33] are also applied.

There are still many gaps in self-awareness research targeting prediction modules. Although some studies have assigned different scores to different trajectories, they were only applicable to the comparison between different outputs of the same prediction task and have limited reference values for reflecting the prediction performance in different scenarios. Some studies model motion uncertainty in the output, such as using variance or covariance [34], [35], but their way of directly optimizing the probability distribution is not conducive to guaranteeing the optimality of the most likely trajectory. The proposed two-stage training method prioritizes the accuracy of the predicted trajectory in the first stage, and the optimization of the self-awareness module in the second stage does not affect the trained trajectory prediction module. Besides, [36] applied the EU quantization techniques to OOD detection in trajectory prediction. [37] explored failure detection for motion prediction models from the perspective of uncertainty.

Different from previous studies, this paper uses supervised learning to guide a lightweight self-awareness module to output diagnostic information with an explicit relationship with prediction error, which improves the accessibility and interpretability of detection results.

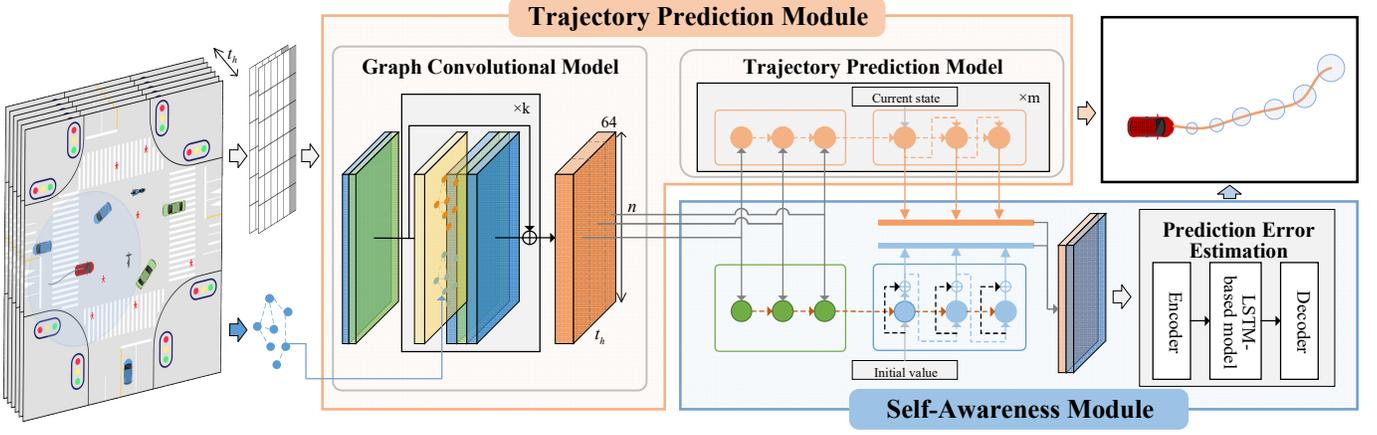

Fig. 2. Overview of our proposed method. The historical trajectories of all observed TPs in the scene are constructed as a ($c \times t_h \times n$)-dimensional tensor, where $c$ is the input feature dimension of each TP at each moment, $n$ is the maximum considered number of TPs. The relationship of different TPs is constructed as a fixed graph. In the self-awareness module, the ($64 \times t_h \times n$)-dimensional middle layer graph feature of the trajectory prediction module and the predicted trajectory are considered as inputs. A separate GRU-based Seq2Seq network is designed to process the graph feature. Then it is fused with the predicted trajectory and fed into the error estimation network, and finally, the diagnostic information for the output of the trajectory prediction module is obtained.

## III. SELF-AWARE TRAJECTORY PREDICTION

In this section, we describe the proposed self-aware trajectory prediction methodology and its implementation. Our approach strives to endow the trajectory prediction model with self-awareness by introducing a module with the ability to estimate the prediction performance online.

### A. Framework Overview

Suppose the current moment $t=0$, given the historical state $\mathbf{X} = [s^{(-t_h+1)}, s^{(-t_h+2)}, \cdots, s^{(0)}]$ of the predicted object over the past $t_h$ moments and the surrounding scene context $\mathbf{C}$, the trajectory prediction module aims to predict its future trajectory $\mathbf{Y} = [s^{(1)}, s^{(2)}, ..., s^{(t_f)}]$, where $t_f$ is the predicted horizon. The motion state $s^{(t)}$ at the $t$-th moment generally includes the object's location, and the scene context $\mathbf{C}$ includes the historical state of surrounding TPs, from which their interactions with the predicted object can be modeled. The predicted trajectory is denoted by $\hat{\mathbf{Y}} = [\hat{s}^{(1)}, \hat{s}^{(2)}, ..., \hat{s}^{(t_f)}]$.

Fig. 2 shows the proposed self-aware trajectory prediction framework. In the trajectory prediction module, based on $\mathbf{X}$ and $\mathbf{C}$, the input tensor is extracted to represent the state of all observed TPs and the fixed graph is constructed to represent the relationship between different TPs. The framework introduces a new self-awareness module based on the trajectory prediction module, which aims to provide a performance diagnosis $\mathbf{Z}$ when the trajectory prediction module outputs the predicted trajectory. It can be a composite value that measures performance on the current prediction task, or a sequence $\hat{\mathbf{Z}} = [\hat{z}^{(1)}, \hat{z}^{(2)}, ..., \hat{z}^{(t_f)}]$ for all future moments. Based on this performance estimation result, IVs can judge whether the current trajectory prediction module is performing well or not, and make safe decisions in scenarios where the trajectory prediction module may fail. The trajectory prediction module is designed by referring to [9]. The following part mainly introduces the design scheme of the self-awareness module.

### B. Self-Awareness Module

*1) Input features:* The input of the self-awareness module is taken from the trajectory prediction module, which mainly contains the graph feature from the middle layer and the predicted trajectory. Some previous research [38], [39] has demonstrated that the middle layer features of the network are valuable for anomaly detection. Specifically, in the trajectory prediction module, a graph feature is constructed through the graph convolutional model, this feature can well represent the interaction in the current scenario, so it is extracted for the self-awareness module. In addition, the self-awareness module aims to diagnose whether the predicted trajectory is valid, so the output $\hat{\mathbf{Y}}$ of the trajectory prediction module is also taken as key information.

*2) Modular architecture:* The proposed self-awareness module consists of three submodules: graph feature processing, feature fusion, and prediction error estimation. The graph feature processing submodule uses a Seq2Seq network to convert the graph feature into a tensor that matches the predicted trajectory's size. This submodule comprises an encoder and decoder based on GRU. In this study, the first decoding step takes a specified value set to 0. The feature fusion submodule combines the processed graph feature with the predicted trajectory using various methods discussed in the experimental section. The prediction error estimation submodule evaluates how well the current trajectory prediction module performs based on fused features. It encodes fused features for each TP at every moment using an encoder, processes temporal relationships through an LSTM-based model, and obtains diagnostic information via a decoder.

*3) Diagnostic information Form:* The self-awareness module should provide diagnostic information that can assess the validity of the predicted trajectory. This study uses the prediction error as the label, which is determined by measuring the distance between predicted and actual trajectory points at each moment in the future:

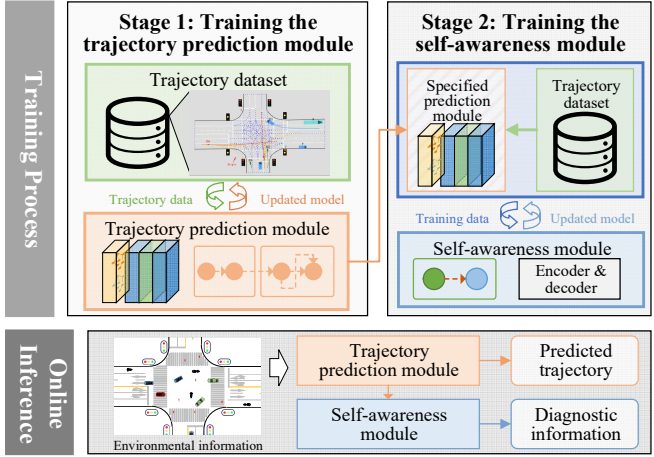

Fig. 3. Overview of the training and inference process. The training process consists of two stages. In stage 1, a trajectory prediction module is trained with trajectory dataset. In stage 2, the self-awareness module is updated based on the trajectory data and the trained trajectory prediction module. In the online inference process, the two modules work simultaneously to output the predicted trajectory and diagnostic information.

$$z^{(t)} = dist\left(s^{(t)}, \hat{s}^{(t)}\right) = \left\| s^{(t)} - \hat{s}^{(t)} \right\|_2, \quad (1)$$

where $t \in \{1, 2, ..., t_f\}$. This form can provide rich information for both the failure detection for the trajectory prediction module and the subsequent safe decision-making.

In addition, the errors of the predicted trajectory in the *x* and *y* coordinates at each time point can also be modeled separately, and the analysis is shown in the experimental part of Section V.

*C. Training and Inference Process*

A specific model development and application process is designed for the proposed self-aware trajectory prediction, as shown in Fig. 3. It should be stated that the introduction of the self-awareness module should not be at the expense of the performance of the original trajectory prediction module. Therefore, the training process is divided into two independent stages: training the trajectory prediction module and training the self-awareness module.

In the first stage, the trajectory prediction module is trained using the trajectory dataset. This allows for the utilization of existing training methods for trajectory prediction. During the actual development process of the IV system, the training of the trajectory prediction module can be handed over to the original developers without adjusting the training process. The training loss in the optimization of the trajectory prediction module is obtained by calculating the error between the true future trajectory and the predicted trajectory.

In the second stage of the training process, the self-awareness module is individually optimized based on the trajectory dataset and the trained trajectory prediction module. To avoid adverse effects on the trajectory prediction module at this stage, we adopt the parameter freezing method to ensure that the parameters of the trajectory prediction module remain unchanged during the process. The training data of this stage is obtained through the trajectory dataset and the prediction module trained in stage 1. The graph feature and predicted trajectories are taken as inputs, and the label is determined according to the form of diagnostic information, as shown in Eq. (2). The optimization at this stage is regarded as a multi-point regression task, and the loss is calculated as follows:

$$L_{sa} = \frac{1}{t_f} \sum_{t=1}^{t_f} \left\| z^t - \hat{z}^t \right\|, \quad (2)$$

The process of decoupling the training of the two modules facilitates the flexible adjustment of the training methods and resources of the two stages in the actual development process. Meanwhile, the self-awareness module can be transferred conveniently between different trajectory prediction modules.

The online inference of the proposed self-aware trajectory prediction model is jointly realized by two modules. Specifically, the trajectory prediction module predicts the TPs' future trajectory based on the environmental information, and it synchronously provides the graph feature and the predicted trajectory to the self-awareness module for self-diagnosis.

*D. Implementation Details*

This study involves a trajectory prediction module that observes 3 seconds of historical trajectory and predicts the next 3 seconds. The trajectory is sampled at 2Hz, which is typical for many trajectory prediction challenges and datasets. The design of the trajectory prediction module follows the scheme in [9]. It uses a graph convolutional model to process input data, including several graph operation layers, temporal 2D convolutional layers, and batch normalization layers. Additionally, a trajectory prediction model composed of three Seq2Seq networks estimates future trajectories. In the self-awareness module, the graph feature processing submodule uses a 2-layer GRU as the basis of the encoder and decoder networks, and the processed graph feature and the predicted trajectory are connected along the dimension of the features at each time. In the error estimation submodule, the multilayer perceptron (MLP) network is used as both the encoder and decoder, and a 2-layer LSTM is used to process the middle layer features. When the Euclidean distance is used as the error label, the ReLU activation function is used to ensure that the output is non-negative. Adam optimizer and StepLR scheduler are used in both training stages. The model is implemented using Python Programming Language and PyTorch library.

IV. EXPERIMENTS

*A. Datasets*

The proposed approach is evaluated on 2 public intersection datasets: SinD [40] and TC_intersection_VA (VA) [41]. The SinD dataset is collected by a drone at a typical signalized intersection and contains 13248 trajectories from vehicles and vulnerable road users (VRUs), such as pedestrians, cyclists, and pedestrians. Besides, many traffic violations are recorded, such as running a red light. These data are divided into 23 records according to the collection time, each with a duration of 8-20 minutes. We select 16 records as the training and validation set, and 7 records as the test set. VA is a subset of the INTERACTION dataset, collected by traffic cameras at an intersection in Bulgaria. It is used as a supplementary test set to evaluate the performance of our method in the presence of some distributional shifts. This dataset contains 60 minutes of data.

*B. Metrics*

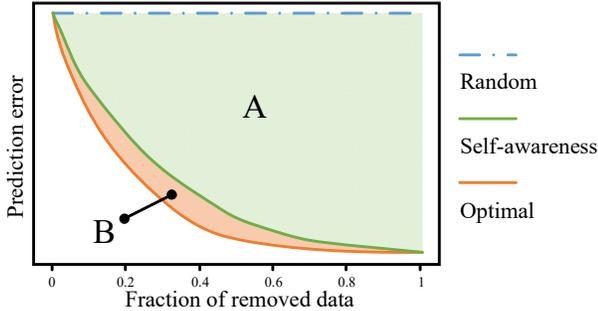

Fig. 4. Overview of the cutoff curve.

Self-aware trajectory prediction is a novel task for which we define some evaluation metrics. We use the cutoff curve [42] to analyze the self-awareness ability of the proposed method, which can reflect the potential of improving the performance of the original system after adding the self-awareness module. As shown in Fig. 4, the test results are sorted in descending order of diagnostic information, and then a certain proportion of data is filtered out in order and the average prediction error of the remaining data is calculated to obtain a point on the cutoff curve. The points for different fractions of removed data together form a cutoff curve. In contrast, the random cutoff curve is obtained by randomly removing the data, and the curve for optimal is obtained by directly taking the true prediction error as the self-aware output. The area under the cutoff curve (AUCOC) represents the overall performance of the self-aware trajectory prediction model and can be used as an evaluation metric.

In addition, to compare different self-awareness modules separately, we propose self-awareness score (SAS) as an important metric:

$$\text{SAS} = \frac{\text{AUCOC}_{\text{random}} - \text{AUCOC}_{\text{self-awareness}}}{\text{AUCOC}_{\text{random}} - \text{AUCOC}_{\text{optimal}}} = \frac{A}{A+B}, \quad (3)$$

where average displacement error (ADE) and final displacement error (FDE) are used as the above prediction errors, respectively. In addition, the point-wise prediction errors output by the self-awareness module are integrated to form the estimates of ADE and FDE, which are used as diagnostic information for the above two types of errors, respectively.

Moreover, autonomous driving is a system with limited computational resources, thus a model with a small memory footprint and fast inference speed is favored. Therefore, we calculate the total number of parameters and the average execution time required for a single frame of the trajectory prediction model with the self-awareness module as an evaluation metric. In each frame, the model predicts the future trajectories of multiple TPs in the scenario and outputs diagnostic information. An NVIDIA GeForce RTX 2080 SUPER GPU and an Inter i7-9700 CPU were used for all experiments to obtain the execution time.

*C. Baseline Methods*

*1) Motion uncertainty (MU) estimation:* As mentioned in [37], we modified the original trajectory prediction network into a model containing a maneuver classification module and calculated the ADE and FDE of the trajectory under the maximum probability maneuver as the prediction error. Two types of indicators are considered diagnostic information. One is the negative maximum softmax probability (NMaP) output by the classification module, as a representative method for the network to directly output the score of each predicted trajectory. The other is average predictive entropy (APE) and final predictive entropy (FPE) calculated based on trajectories for different maneuvers, where APE and FPE are used as diagnostic information for ADE and FDE, respectively.

*2) EU estimation:* Two typical EU estimation methods were implemented as baselines: MC dropout and deep ensemble. For MC dropout, after the model is trained based on the specific strategy [17], the dropout remains on during the inference process, and multiple samples are taken to obtain multiple predictions, where the dropout rate is set to 0.5 and the sampling number is set to 5. For deep ensemble, 5 submodels are trained by random initialization and random shuffling, then the results of all submodels are obtained simultaneously in the inference process. Based on all the predictions above, an average trajectory is obtained to calculate ADE and FDE, APE and FPE from all trajectories are calculated as the diagnostic information.

*3) AE-based method:* Based on the original prediction model, a new decoder module is added to decode the graph feature into historical trajectories, so that the reconstruction errors can be calculated as diagnostic information.

## V. RESULTS AND DISCUSSION

*A. Comparison of Different Methods*

On the SinD and VA datasets, we compare the proposed approach with the baselines described above. TABLE I demonstrates the results of the different methods. Among all baselines, the deep ensemble-based method can achieve the smallest AUCOC and the best SAS, but due to the need to save the parameters of multiple submodels and obtain the results of all submodels simultaneously during execution, its memory footprint and time cost are much higher than those of single-model-based methods, such as AE-based method. Comparing the two error indicators on the two datasets, the proposed method achieves better results in terms of self-awareness than the deep ensemble-based method. Moreover, since the self-awareness module is constituted by a lightweight model, it also achieves good results in terms of the total number of parameters and the average execution time, which shows the good potential of the proposed method for practical applications.

*B. Ablation Study*

A comprehensive ablation study is performed to compare the effects of different input forms, models, and output forms of the self-awareness module. In addition, we compared the results of the proposed two-stage training process with the single-stage training process method in the experiment.

*1) Comparison of different input forms & key models:* Three different input forms are considered according to whether the

TABLE I. COMPARISON OF DIFFERENT METHODS. THE METRICS FOR COLUMNS 2-5 ARE AUCOC (M) (↓)/SAS(↑).

| | SinD | | VA | | Total parameters | Average Execution Time (ms/frame) |
|---|---|---|---|---|---|---|
| | ADE | FDE | ADE | FDE | | |
| MU(NMap-based) | 0.129/0.698 | 0.280/0.688 | 0.381/0.535 | 0.862/0.520 | 572966 | 23.69 |
| MU(APE/FPE-based) | 0.111/0.766 | 0.261/0.720 | 0.305/0.683 | 0.709/0.648 | 572966 | 23.69 |
| EU(MC dropout) | 0.088/0.869 | 0.189/0.856 | 0.334/0.686 | 0.742/0.656 | 496302 | 36.74 |
| EU(deep ensemble) | 0.062/0.893 | 0.142/0.862 | 0.252/0.737 | 0.579/0.704 | 2481510 | 36.75 |
| AE-based | 0.083/0.826 | 0.179/0.813 | 0.283/0.686 | 0.636/0.666 | 615545 | 11.31 |
| **Ours** | **0.061/0.919** | **0.130/0.908** | **0.245/0.763** | **0.552/0.738** | 615645 | **10.66** |

TABLE II. SAS(↑) FOR DIFFERENT INPUT FORMS & BASIC MODELS OF THE PREDICTION ERROR ESTIMATION SUBMODULE.

| Input | basic model | SinD | | VA | |
|---|---|---|---|---|---|
| | | ADE | FDE | ADE | FDE |
| GF | - | 0.886 | 0.876 | 0.744 | 0.724 |
| GF | MLP | 0.901 | 0.891 | 0.747 | 0.723 |
| GF | Conv | 0.903 | 0.893 | 0.754 | 0.728 |
| GF | LSTM | 0.902 | 0.890 | 0.752 | 0.725 |
| add | MLP | 0.886 | 0.874 | 0.750 | 0.726 |
| add | Conv | 0.887 | 0.875 | 0.750 | 0.722 |
| add | LSTM | 0.896 | 0.884 | 0.763 | 0.738 |
| concat | MLP | 0.889 | 0.879 | 0.745 | 0.718 |
| concat | Conv | 0.899 | 0.889 | 0.755 | 0.728 |
| **concat** | **LSTM (ours)** | **0.919** | **0.908** | **0.763** | **0.738** |

TABLE III. SAS(↑) FOR DIFFERENT OUTPUT FORMS OF THE SELF-AWARENESS MODULE.

| Error label | SinD | | VA | |
|---|---|---|---|---|
| | ADE | FDE | ADE | FDE |
| Velocity | 0.814 | 0.773 | 0.707 | 0.670 |
| Position | 0.837 | 0.785 | 0.721 | 0.699 |
| **Distance (ours)** | **0.919** | **0.908** | **0.763** | **0.738** |

self-awareness module takes the predicted trajectory as input and how the features are fused: graph feature only (GF), the processed graph feature is added (add), or concatenated (concat) with the predicted trajectory. In addition, we compare the SAS when using MLP, 1D convolution (Conv), or LSTM as the basic model of the prediction error estimation submodule or without the submodule. The experimental results are recorded in TABLE II. It is observed that the introduction of the prediction error estimation submodule effectively improves the ability of the self-awareness module. Although the method of only taking graph feature as input has achieved a good result, better performance can be achieved by fusing the predicted trajectories in a proper way, where the concatenation operation is considered to be superior to the addition operation. From the comparison of different basic models, it is found that LSTM provides more effective support than MLP and Conv in most cases, which may be due to its advantages in processing time series data. In summary, the proposed architecture achieves the best results in different datasets and prediction error metrics.

*2) Comparison of different output forms:* As described in Section III, different types of diagnostic information correspond to different forms of self-awareness module output. This section studies the effects of different output forms, as shown in TABLE III. Three output forms are compared, that is, estimates for all future moments: the error of the respective predicted speed in the *x* and *y* directions, the error of the respective predicted position in the *x* and *y* directions, and the error directly measured by the Euclidean distance. Correspondingly, the error labels used in training are also adjusted according to the output form. Compared with the third form, the first two forms above distinguish the errors in the *x* and *y* directions, but the results show that the relatively simplified output form leads to a significant improvement in the ability of the self-awareness module. We believe that the acceptable simplification of the diagnostic information helps guide the self-awareness module to focus on learning key patterns while increasing its robustness to adverse factors such as noise.

*3) Comparison of different training processes:* The proposed two-stage training process avoid adverse effects on the original trajectory prediction module during the optimization process of the self-awareness module. TABLE IV demonstrates the results of different training processes. The third row represents the results of optimizing the trajectory prediction and self-awareness modules at the same time by weighting different losses. The loss function is set as follows:

$$L = L_{tp} + \lambda L_{sa}, \quad (4)$$

where $\lambda$ is the weight coefficient, which is set to 0.1.

The results show that although synchronous training can realize the optimization of the self-awareness module, it may lead to the degradation of the performance of the original trajectory prediction module, and thus lead to the increase of AUCOC. In addition, the single-stage training method has strict requirements on the weight coefficient, lacks flexibility, and lacks applicability when it is difficult to obtain the adjustment right of the original trajectory prediction module parameters. Therefore, the proposed two-stage training method is valuable.

TABLE IV. AUCOC (M)(↓)/SAS(↑) FOR DIFFERENT TRAINING PROCESSES.

| Training process | SinD | | VA | |
|---|---|---|---|---|
| | ADE | FDE | ADE | FDE |
| weighting | 0.065/0.916 | 0.138/0.904 | 0.255/0.762 | 0.579/0.735 |
| **Ours** | **0.061/0.919** | **0.130/0.908** | **0.245/0.763** | **0.552/0.738** |

## C. Other Quantitative Analysis

*1) Comparison of results at different moments:* To further analyze the performance of the proposed method, we separately explore the diagnostic ability of the self-awareness module for the trajectory prediction module at each moment in the future. Specifically, the distance between the predicted position and the real position at each moment is regarded as the prediction error used to draw the cutoff curve, and the estimated error at the corresponding moment output by the self-awareness module is regarded as diagnostic information. The results are shown in TABLE V. It is observed that SASs at different moments maintains a relatively close level, and the SAS for the nearest moment (0.5s) in the future is relatively low. One of the possible reasons is that the original predicted position at this moment has already had relatively high accuracy, so the performance

TABLE V. Results for Different Moments in the Future: $AUCOC_{RANDOM}(M)/AUCOC_{SELF-AWARENESS}(M)(\downarrow)/AUCOC_{OPTIMAL}(M)/SAS(\uparrow)$.

| Moment | SinD | VA |
|---|---|---|
| 1 | 0.040/0.013/0.008/0.819 | 0.069/0.036/0.021/0.695 |
| 2 | 0.099/0.024/0.017/0.918 | 0.198/0.088/0.052/0.757 |
| 3 | 0.184/0.043/0.030/0.920 | 0.389/0.158/0.089/0.772 |
| 4 | 0.294/0.066/0.046/0.918 | 0.645/0.255/0.135/0.765 |
| 5 | 0.427/0.066/0.046/0.914 | 0.973/0.385/0.192/0.753 |
| 6 | 0.580/0.130/0.084/0.908 | 1.372/0.552/0.261/0.738 |

improvement brought by the self-awareness module is limited due to the long tail problem.

*2) Comparison of results of different types of TPs:* Intersection scenarios include various types of TPs, with varying behavioral patterns and prediction difficulties. The results for different types of TPs are recorded in TABLE VI. It is observed that the self-awareness performance for trajectory prediction for pedestrians and non-motorized vehicles is not as good as that of motor vehicles. At signalized intersections, the movement of motor vehicles is generally regular and predictable. In contrast, the movement of pedestrians and non-motorized vehicles is more random, accompanied by a large number of unpredictable violations, which poses challenges to both trajectory prediction and self-awareness modules.

TABLE VI. SAS($\uparrow$) for Different Types of TPs.

| Type | SinD | | VA | |
|---|---|---|---|---|
| | ADE | FDE | ADE | FDE |
| Small vehicle | 0.916 | 0.901 | 0.762 | 0.737 |
| Big vehicle | 0.973 | 0.977 | 0.775 | 0.769 |
| Pedestrian | 0.799 | 0.772 | - | - |
| Motorcyclist and bicyclist | 0.878 | 0.866 | - | - |

*D. Qualitative Analysis*

Fig. 5 visualizes the results of the proposed method. It is observed that the original trajectory prediction module underperform in some challenging scenarios, such as turning, complex interaction or movement. The introduction of the self-awareness module helps to estimate the poor prediction performance in these scenarios online and increases the usability of the prediction results by complementing the prediction area.

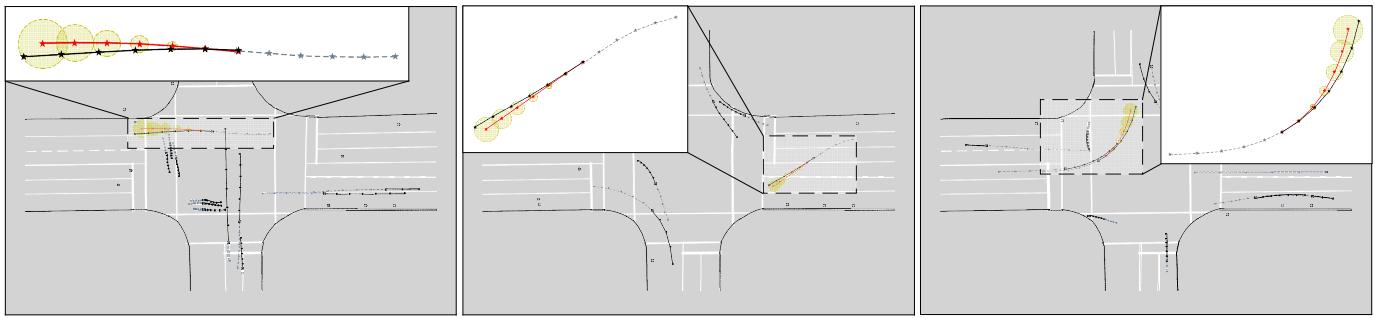

Fig. 5. Visualizations. The slate gray dotted line represents the historical trajectory, the black solid line represents the real future trajectory, the red solid line represents the predicted trajectory, and the yellow area is formed based on the distance error estimated by the self-awareness module.

## VI. CONCLUSION

In this work, we propose a self-aware trajectory prediction framework that enables IVs to self-diagnose their predictions while predicting the trajectories of surrounding TPs, thereby facilitating risk mitigation for potential triggering scenarios where trajectory prediction fails. A novel self-awareness module is introduced into the existing trajectory prediction system to estimate the prediction performance. A two-stage training process is designed to optimize the self-awareness module without affecting the performance of the original trajectory prediction module. A comprehensive comparison experiment is performed, and the proposed method achieves valuable results in the evaluation of self-awareness ability and real-time performance. Moreover, further quantitative and qualitative analyzes of the proposed method reveal more details.

This work provides a paradigm worthy of reference for building reliable AI-based AD software. It has good application potential for other forms of prediction, but more details are worth exploring, such as self-awareness for multimodal trajectory prediction. In addition, this work mainly focuses on the online diagnosis of the trajectory prediction module, our future work will explore combining the proposed method with safe decision-making to effectively improve IV's safety. For example, the diagnostic results from the self-awareness module can be used as additional information to IV's tactical planner, providing more reliable predictions for other TPs.

REFERENCES

[1] Y. Huang, J. Du, Z. Yang, Z. Zhou, L. Zhang, and H. Chen, "A Survey on Trajectory-Prediction Methods for Autonomous Driving," *IEEE Transactions on Intelligent Vehicles*, vol. 7, no. 3, pp. 652–674, Sep. 2022, doi: 10.1109/TIV.2022.3167103.

[2] N. Deo, E. Wolff, and O. Beijbom, "Multimodal Trajectory Prediction Conditioned on Lane-Graph Traversals," in *Proceedings of the 5th Conference on Robot Learning*, PMLR, Jan. 2022, pp. 203–212.

[3] Q. Sun, X. Huang, J. Gu, B. C. Williams, and H. Zhao, "M2I: From Factored Marginal Trajectory Prediction to Interactive Prediction," presented at the Proceedings of the IEEE/CVF Conference on Computer Vision and Pattern Recognition, 2022, pp. 6543–6552.

[4] Board, TNTS, "Highway Accident Report: Collision Between Vehicle Controlled by Developmental Automated Driving System and Pedestrian," Nov. 2019. Accessed: Dec. 20, 2022. [Online]. Available: https://trid.trb.org/view/1751168

[5] J. Schlatow *et al.*, "Self-awareness in autonomous automotive systems," in *Design, Automation & Test in Europe Conference & Exhibition (DATE), 2017*, Mar. 2017, pp. 1050–1055. doi: 10.23919/DATE.2017.7927145.

[6] L. A. Dennis and M. Fisher, "Verifiable Self-Aware Agent-Based Autonomous Systems," *Proceedings of the IEEE*, vol. 108, no. 7, pp. 1011–1026, Jul. 2020, doi: 10.1109/JPROC.2020.2991262.

[7] A. Alahi, K. Goel, V. Ramanathan, A. Robicquet, L. Fei-Fei, and S. Savarese, "Social LSTM: Human Trajectory Prediction in Crowded Spaces," in *Proceedings of the IEEE Conference on Computer Vision and Pattern Recognition*, 2016, pp. 961–971.

[8] Y. Liu, J. Zhang, L. Fang, Q. Jiang, and B. Zhou, "Multimodal Motion Prediction With Stacked Transformers," presented at the Proceedings of the IEEE/CVF Conference on Computer Vision and Pattern Recognition, 2021, pp. 7577–7586.

[9] X. Li, X. Ying, and M. C. Chuah, "GRIP++: Enhanced Graph-based Interaction-aware Trajectory Prediction for Autonomous Driving." arXiv, May 19, 2020. doi: 10.48550/arXiv.1907.07792.

[10] J. Gao *et al.*, "VectorNet: Encoding HD Maps and Agent Dynamics From Vectorized Representation," in *Proceedings of the IEEE/CVF Conference on Computer Vision and Pattern Recognition*, 2020, pp. 11525–11533.

[11] M. Liang *et al.*, "Learning Lane Graph Representations for Motion Forecasting," in *Computer Vision – ECCV 2020*, A. Vedaldi, H. Bischof, T. Brox, and J.-M. Frahm, Eds., in Lecture Notes in Computer Science. Cham: Springer International Publishing, 2020, pp. 541–556. doi: 10.1007/978-3-030-58536-5_32.

[12] H. Zhao *et al.*, "TNT: Target-driven Trajectory Prediction," in *Proceedings of the 2020 Conference on Robot Learning*, PMLR, Oct. 2021, pp. 895–904.

[13] J. Gu, C. Sun, and H. Zhao, "DenseTNT: End-to-End Trajectory Prediction From Dense Goal Sets," in *Proceedings of the IEEE/CVF International Conference on Computer Vision*, 2021, pp. 15303–15312.

[14] T. Salzmann, B. Ivanovic, P. Chakravarty, and M. Pavone, "Trajectron++: Dynamically-Feasible Trajectory Forecasting with Heterogeneous Data," in *Computer Vision – ECCV 2020*, A. Vedaldi, H. Bischof, T. Brox, and J.-M. Frahm, Eds., in Lecture Notes in Computer Science. Cham: Springer International Publishing, 2020, pp. 683–700. doi: 10.1007/978-3-030-58523-5_40.

[15] J. Gawlikowski *et al.*, "A Survey of Uncertainty in Deep Neural Networks." arXiv, Jan. 18, 2022. doi: 10.48550/arXiv.2107.03342.

[16] C. Blundell, J. Cornebise, K. Kavukcuoglu, and D. Wierstra, "Weight Uncertainty in Neural Network," in *Proceedings of the 32nd International Conference on Machine Learning*, PMLR, Jun. 2015, pp. 1613–1622.

[17] Y. Gal and Z. Ghahramani, "Dropout as a Bayesian Approximation: Representing Model Uncertainty in Deep Learning," in *Proceedings of The 33rd International Conference on Machine Learning*, PMLR, Jun. 2016, pp. 1050–1059.

[18] Y. Gal, J. Hron, and A. Kendall, "Concrete Dropout," in *Advances in Neural Information Processing Systems*, Curran Associates, Inc., 2017.

[19] B. Lakshminarayanan, A. Pritzel, and C. Blundell, "Simple and Scalable Predictive Uncertainty Estimation using Deep Ensembles," in *Advances in Neural Information Processing Systems*, Curran Associates, Inc., 2017.

[20] Y. Wen *et al.*, "Combining Ensembles and Data Augmentation can Harm your Calibration." arXiv, Mar. 22, 2021. doi: 10.48550/arXiv.2010.09875.

[21] L. Ruff *et al.*, "A Unifying Review of Deep and Shallow Anomaly Detection," *Proceedings of the IEEE*, vol. 109, no. 5, pp. 756–795, 2021, doi: 10/gjmk3g.

[22] D. Bogdoll, M. Nitsche, and J. M. Zöllner, "Anomaly Detection in Autonomous Driving: A Survey," in *Proceedings of the IEEE/CVF Conference on Computer Vision and Pattern Recognition*, 2022, pp. 4488–4499.

[23] A. Borghesi, A. Bartolini, M. Lombardi, M. Milano, and L. Benini, "Anomaly Detection Using Autoencoders in High Performance Computing Systems," *Proceedings of the AAAI Conference on Artificial Intelligence*, vol. 33, no. 01, pp. 9428–9433, Jul. 2019, doi: 10.1609/aaai.v33i01.33019428.

[24] P. Cui and J. Wang, "Out-of-Distribution (OOD) Detection Based on Deep Learning: A Review," *Electronics*, vol. 11, no. 21, p. 3500, Jan. 2022, doi: 10.3390/electronics11213500.

[25] D. Hendrycks and K. Gimpel, "A Baseline for Detecting Misclassified and Out-of-Distribution Examples in Neural Networks," in *International Conference on Learning Representations*, 2017.

[26] J. Guérin, K. Delmas, R. S. Ferreira, and J. Guiochet, "Out-Of-Distribution Detection Is Not All You Need." arXiv, Nov. 29, 2022.

[27] J. Hawke, H. E, V. Badrinarayanan, and A. Kendall, "Reimagining an autonomous vehicle." arXiv, Aug. 12, 2021. doi: 10.48550/arXiv.2108.05805.

[28] C. B. Kuhn, M. Hofbauer, G. Petrovic, and E. Steinbach, "Introspective Black Box Failure Prediction for Autonomous Driving," in *2020 IEEE Intelligent Vehicles Symposium (IV)*, Oct. 2020, pp. 1907–1913. doi: 10.1109/IV47402.2020.9304844.

[29] M. S. Ramanagopal, C. Anderson, R. Vasudevan, and M. Johnson-Roberson, "Failing to Learn: Autonomously Identifying Perception Failures for Self-Driving Cars," *IEEE Robotics and Automation Letters*, vol. 3, no. 4, pp. 3860–3867, Oct. 2018, doi: 10.1109/LRA.2018.2857402.

[30] L. Biddle and S. Fallah, "A Novel Fault Detection, Identification and Prediction Approach for Autonomous Vehicle Controllers Using SVM," *Automot. Innov.*, vol. 4, no. 3, pp. 301–314, Aug. 2021, doi: 10.1007/s42154-021-00138-0.

[31] B. Sun, J. Xing, H. Blum, R. Siegwart, and C. Cadena, "See Yourself in Others: Attending Multiple Tasks for Own Failure Detection," in *2022 International Conference on Robotics and Automation (ICRA)*, 2022, pp. 8409–8416. doi: 10.1109/ICRA46639.2022.9812310.

[32] M. Rottmann *et al.*, "Prediction Error Meta Classification in Semantic Segmentation: Detection via Aggregated Dispersion Measures of Softmax Probabilities," in *2020 International Joint Conference on Neural Networks (IJCNN)*, Jul. 2020, pp. 1–9. doi: 10.1109/IJCNN48605.2020.9206659.

[33] Di Biase G., Blum H., Siegwart R., and Cadena C., "Pixel-Wise Anomaly Detection in Complex Driving Scenes," presented at the Proceedings of the IEEE/CVF Conference on Computer Vision and Pattern Recognition, 2021, pp. 16918–16927.

[34] J. Strohbeck *et al.*, "Multiple Trajectory Prediction with Deep Temporal and Spatial Convolutional Neural Networks," in *2020 IEEE/RSJ International Conference on Intelligent Robots and Systems (IROS)*, Oct. 2020, pp. 1992–1998. doi: 10.1109/IROS45743.2020.9341327.

[35] J. Strohbeck, J. Müller, M. Herrmann, and M. Buchholz, "Deep Kernel Learning for Uncertainty Estimation in Multiple Trajectory Prediction Networks," in *2022 IEEE/RSJ International Conference on Intelligent Robots and Systems (IROS)*, Oct. 2022, pp. 11396–11402. doi: 10.1109/IROS47612.2022.9982167.

[36] M. Itkina and M. J. Kochenderfer, "Interpretable Self-Aware Neural Networks for Robust Trajectory Prediction." arXiv, Nov. 16, 2022. doi: 10.48550/arXiv.2211.08701.

[37] W. Shao, Y. Xu, L. Peng, J. Li, and H. Wang, "Failure Detection for Motion Prediction of Autonomous Driving: An Uncertainty Perspective." arXiv, Jan. 11, 2023. doi: 10.48550/arXiv.2301.04421.

[38] S. Luan, Z. Gu, L. B. Freidovich, L. Jiang, and Q. Zhao, "Out-of-Distribution Detection for Deep Neural Networks With Isolation Forest and Local Outlier Factor," *IEEE Access*, vol. 9, pp. 132980–132989, 2021, doi: 10.1109/ACCESS.2021.3108451.

[39] F. Zhao, C. Zhang, N. Dong, Z. You, and Z. Wu, "A Uniform Framework for Anomaly Detection in Deep Neural Networks," *Neural Process Lett*, vol. 54, no. 4, pp. 3467–3488, Aug. 2022, doi: 10.1007/s11063-022-10776-y.

[40] Y. Xu *et al.*, "SIND: A Drone Dataset at Signalized Intersection in China." arXiv, Sep. 06, 2022. doi: 10.48550/arXiv.2209.02297.

[41] W. Zhan *et al.*, "INTERACTION Dataset: An INTERnational, Adversarial and Cooperative moTION Dataset in Interactive Driving Scenarios with Semantic Maps." arXiv, Sep. 30, 2019. doi: 10.48550/arXiv.1910.03088.

[42] J. Lindqvist, A. Olmin, F. Lindsten, and L. Svensson, "A general framework for ensemble distribution distillation," in *2020 IEEE 30th International Workshop on Machine Learning for Signal Processing (MLSP)*, IEEE, 2020, pp. 1–6. doi: 10/gh9szc.